\documentclass{article}
\usepackage[T1]{fontenc}
\usepackage{iclr2027_conference,times}
\usepackage{amsmath,amssymb}
\usepackage{graphicx}
\usepackage{hyperref}
\usepackage{url}
\usepackage{wrapfig}
\usepackage{float}
\usepackage{booktabs}
\usepackage{fvextra}
\usepackage{upquote}
\usepackage{caption}
\usepackage[most]{tcolorbox}
\usepackage{colortbl}
\usetikzlibrary{positioning,arrows.meta}
\definecolor{findingrule}{HTML}{2A5DB0}
\definecolor{citecolor}{HTML}{0071BC}
\definecolor{codeurlcolor}{cmyk}{0,1,0,0}
\definecolor{findingorangebg}{HTML}{FCF1E7}
\definecolor{findingorange}{HTML}{C2611F}
\definecolor{codebackground}{HTML}{F7F7F7}
\definecolor{codeforeground}{HTML}{1F2328}
\definecolor{codenumber}{HTML}{5F6B66}
\newcommand{\shade}[2]{\cellcolor{findingorange!#1}#2}
\hypersetup{colorlinks=true,citecolor=citecolor,linkcolor=black,urlcolor=black}
\newtcolorbox{findingbox}{enhanced,breakable=false,frame hidden,colback=findingorangebg,
  borderline west={2.2pt}{0pt}{findingorange},sharp corners,
  left=9pt,right=9pt,top=5pt,bottom=5pt,before skip=10pt,after skip=8pt}
\newcommand{\finding}[2]{\begin{findingbox}{\color{findingorange}\textsc{\textbf{Finding #1}}}\quad\textit{#2}\end{findingbox}}

\title{Environment Duels:\\A Microcosm of LLM Co-Evolution}
\iclrfinalcopy
\author{%
Shibo Jin\thanks{\fontsize{8.8}{10}\selectfont\mbox{Equal contribution, Corresponding Authors:
\texttt{\{shibo517,oscarxzq\}@engineering.upenn.edu}.}}\quad
Zhiqiu Xu\footnotemark[1]\quad
Yiling Zou\quad
Shangyu Gong\quad
Mayur Naik\\
University of Pennsylvania
}
\hypersetup{%
  pdftitle={Environment Duels: A Microcosm of LLM Co-Evolution},%
  pdfauthor={Shibo Jin, Zhiqiu Xu, Yiling Zou, Shangyu Gong, Mayur Naik}%
}

\begin{document}
\maketitle
\vspace{-22pt}
\lhead{Preprint}
\begingroup
\setlength{\parskip}{0pt}
\begin{abstract}
Through a tangle of real-world interactions,
from synthetic training pipelines to data shared on the open web,
frontier LLMs absorb what their predecessors and competitors create.
How well does each model play its role in this ecosystem,
as a creator and as a solver?
We introduce \textbf{Environment Duels},
a dynamic agentic benchmark that evaluates both capabilities on interactive environments.
Each model writes environments,
each with a privileged hint,
aiming for tasks that capable solvers fail without the hint and pass with it.
Across nine frontier LLMs,
solving and authoring prove to be distinct skills:
the strongest solvers are not the strongest authors,
while smaller models learn more from how others played
and write the hardest environments and the most useful hints.
Beyond the scores,
the pairwise interactions form a microcosm of how models affect one another.
Head-to-head outcomes are transitive and so yield a stable ranking;
independent authors converge on the same designs;
and in a reinforcement-learning study with \texttt{Qwen3.8-27B},
the learner's hint lift on an author's environments tracks how much training on those environments improves it on held-out reasoning and agentic benchmarks,
whether the environments are its own or another model's.
Each of these patterns has a counterpart in how models shape one another outside the duel.
By staging these interactions under controlled conditions,
Environment Duels offers a window onto how LLMs teach,
challenge and learn from one another.
Code is available at {\hypersetup{urlcolor=codeurlcolor}\url{https://github.com/LLM-Duels/EnvDuels}}.
\end{abstract}
\endgroup
\section{Introduction}
Every frontier LLM is now,
in part,
a student of its predecessors and competitors.
Model developers describe pipelines in which one model authors coding challenges,
math problems or reasoning environments,
and the next model trains on them \citep{xu2026deepseek,zhao2026absolute,cai2025autoforge}.
The same synthetic data spreads through public repositories and benchmarks,
where other labs' training runs ingest it~\citep{drayson2025machine}.
Models thus co-evolve:
each one's training distribution is shaped by what other models chose to write.
When a lab closes this loop deliberately,
having its own models write the curriculum for the next generation,
the process becomes recursive self-improvement in a concrete engineering sense.
Either way,
what a model learns depends on how well another model can write problems worth learning from~\citep{xu2025stronger}.

A model's skill as an author therefore shapes what the next generation learns,
in its own family or a competitor's.
Language models now match or exceed human performance on many established benchmarks~\citep{center2026benchmark}.
As they do,
sufficiently challenging human-authored tasks become increasingly costly to curate~\citep{henniger2026tokengamesevaluatinglanguage},
and recent progress on long-standing open problems,
including Millennium Prize conjectures~\citep{openai2026navierstokes},
suggests that even the supply of unsolved challenges a model can train on is finite.
Once the stock of existing problems thins out,
the bottleneck moves: what matters is the ability to pose new problems that are worth solving.
A model that can formulate well-specified,
novel challenges,
and calibrate their difficulty to push the boundaries of current methods,
is the kind of capability that could drive fully autonomous scientific research.

Environment Duels builds a microcosm of this ecosystem.
Every model plays two roles:
it writes executable environments for the others to solve,
and it solves the environments they write.
An environment is an interactive program with hidden state,
a reward and an action budget,
a richer format than a static text question:
solvers must explore,
act and adapt over many turns rather than answer once.
Each author also supplies a privileged hint that lets a solver approach its environment correctly,
and aims for an environment that capable solvers fail without the hint and pass with it.
After the first round,
each author sees how the others actually played and answers with one counter-move.
We run this duel among nine frontier models,
90 environments and 12{,}960 episodes in all,
and score each model on how well it solves (\textit{solve rate}),
how hard its environments are (\textit{difficulty}),
and how much its hints help (\textit{hint lift}).

What emerges resembles the ecosystem outside the duel.
The models that solve best are not the ones whose environments teach best:
smaller models learn more from how others played,
and their counter-moves produce the hardest environments and the most useful hints.
\texttt{DeepSeek-V4.1-Flash},
for example,
writes the hardest environments of all nine models,
and its developers report training it heavily on synthetic environments \citep{xu2026deepseek},
an independent sign that the duel measures a skill that training can build.
The duel also carries signal beyond itself:
how much an author's hints help a learner anticipates how much that learner gains from training on the author's environments.

Our contributions are as follows.
\begin{enumerate}
\setlength{\itemsep}{1pt}
\item Environment Duels is an agentic benchmark,
and a source of RL environments,
that refreshes itself:
because the models under evaluation write every environment,
its ceiling rises with the strongest author at the cost of tokens rather than human annotation.
\item Nine frontier models write,
solve and respond to one another's environments,
and the resulting duel dynamics reflect how the wider model ecosystem behaves outside the duel.
\item A reinforcement-learning experiment shows that a learner's hint lift on each author's environments tracks how much training on them improves it on held-out benchmarks.
\item All environments,
the pipeline that generates them,
the solver trajectories and the analysis code will be released,
and Environment Duels will be updated as new models enter.
\end{enumerate}

\section{Methodology}
\begin{figure}[t]
\centering
\includegraphics[width=\textwidth]{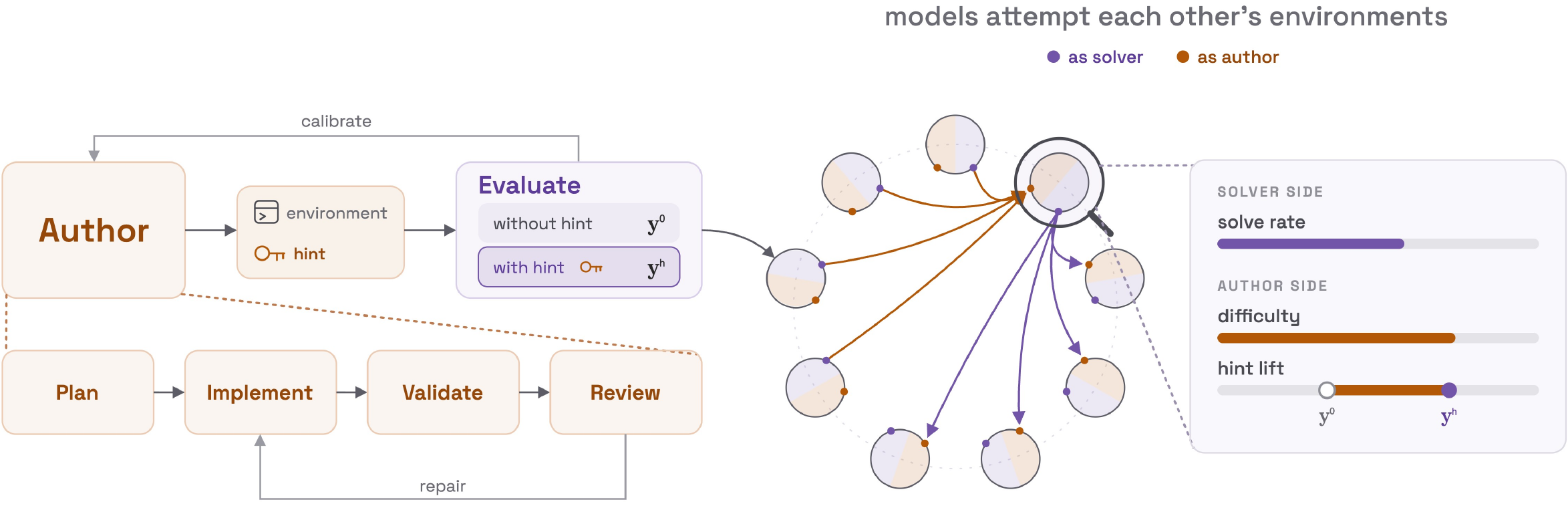}
\caption{\textbf{Overview of Environment Duels.}
Frontier LLMs create and solve each other's environments.
These interactions form a microcosm of the real-world model ecosystem.}
\label{fig:pipeline}
\end{figure}

In Environment Duels,
every evaluated model is both an author and a solver.
A designer writes an executable environment together with a privileged hint; every other model then plays it twice on identical instances,
once with the hint in its context and once without.
Measured outcomes then return to the designer,
which answers with one counter-move,
a revised environment shaped by how the other models played.
Figure~\ref{fig:pipeline} summarises the framework.
We describe construction (Section~\ref{sec:construct}),
then the duel and its scores (Section~\ref{sec:duel}).

\subsection{Constructing an environment}
\label{sec:construct}
Each designer is assigned a domain from a fixed set of five
(symbolic reasoning,
planning and optimization,
causal diagnosis,
adversarial strategy,
dynamic control).
The domain determines the mechanism the environment must exercise.
Within that domain the designer authors an environment that capable solvers should fail without the hint and pass with it,
working through four stages:

\begin{center}
\small
\begin{tikzpicture}[
  stage/.style={draw, rectangle, inner sep=3pt, text height=1.6ex, text depth=.4ex},
  every edge/.style={draw, -{Latex[length=1.6mm]}},
  node distance=6mm]
\node[stage] (plan) {Plan};
\node[stage, right=of plan] (impl) {Implement};
\node[stage, right=of impl] (val) {Validate};
\node[stage, right=of val] (rev) {Review};
\node[right=of rev] (adm) {\emph{admitted}};
\path (plan) edge (impl) (impl) edge (val) (val) edge (rev) (rev) edge (adm);
\draw[-{Latex[length=1.6mm]}] (rev.south) -- ++(0,-3.5mm) -| (plan.south)
  node[pos=0.25, below, font=\footnotesize] {$\leq$2$\times$ repair};
\end{tikzpicture}
\end{center}

\paragraph{Plan.} Before writing any code,
the author produces a design specification,
the exact hint that solvers will receive,
and a private feasibility analysis:
a proposed hinted strategy with turn and resource accounting.
This analysis is written reasoning rather than an executable action sequence,
and neither it nor the plan is ever shown to solvers.
A dedicated planning call reserves the author's full reasoning budget for task design:
working out the core difficulty,
the conditions under which the hint resolves it,
and an adversarial check against simple bypass strategies.
This keeps the design effort separate from implementation effort,
so that the environments are harder and better calibrated.
\paragraph{Implement.} A second call receives the locked plan and hint and writes the environment source,
which must not reveal the hint or give away its strategic insight in any observation.
Formally,
an environment $e=(\mathcal{X},\mathcal{O},\mathcal{A},\mathrm{reset},\mathrm{step},T)$ has hidden states $\mathcal{X}$,
text observations $\mathcal{O}$ and actions $\mathcal{A}$,
and an action budget $T$.
A seed $s$ fixes the instance,
each action advances it deterministically,
and a play succeeds when the environment ends it with positive reward within the budget:
\[
\mathrm{reset}(s)=(x_0,o_0),
\quad
\mathrm{step}(x_t,a_t)=(x_{t+1},o_{t+1},r_{t+1},d_{t+1}),
\quad
y=\mathbf{1}[\tau\le T,\ r_\tau>0],
\]
with reward $r$,
termination flag $d$ and $\tau$ the first step with $d_\tau=1$.
The solver never sees $x_t$ and acts on its history $(o_0,a_0,\dots,o_t)$,
preceded by the hint $h$ in the hinted condition.
In code,
$e$ is a single Python class exposing \texttt{reset(seed)} and \texttt{step(action)},
restricted to the standard library,
with deterministic replay and $T=24$.
The implementation call focuses entirely on translating the plan into correct code;
all design decisions have already been made.

\paragraph{Validate and review.} The pair $(e,h)$ must pass both a programmatic execution check and a self-review by the author in a fresh session before $e$ is admitted to the duel.
A programmatic execution check runs $e$ in an isolated sandbox
(no file-system or network access, fixed time and memory limits),
resets it on two seeds and plays a few probe actions to confirm that observations,
rewards and termination flags are well formed.
For the self-review, the author examines the anonymised $(e,h)$ in a fresh session for reachability within $T$,
instruction consistency,
information sufficiency and hint correctness. If the self-review rejects $(e,h)$ as invalid,
the workflow returns to planning:
the author revises the plan and hint before generating a fresh implementation.
After the duel runs,
we audit environment--seed instances for which no evaluated model succeeds in at least one hint condition.
Each flagged instance undergoes a white-box LLM audit with access to the environment source code and solver trajectories,
and a separate manual review.
We exclude flagged instances judged invalid by both audits from downstream metrics;
the other seeds of the same environment remain eligible.
Appendix~\ref{app:verification} reports the audit procedure and results.

\paragraph{Repair.} A failed execution check returns the error to the implementation call.
A review rejection returns to planning instead:
the author receives the verdict and the failing source,
revises the plan and hint,
and a fresh implementation is generated from the updated plan.
Each environment gets two such repairs.
If both fail,
the author discards the design and starts again from a new plan,
at most twice,
and an environment that still fails is dropped.
In practice the budget is rarely exhausted.
The full prompts for each stage are given in Appendix~\ref{app:prompts}.

\subsection{The duel}
\label{sec:duel}
The duel measures how well each model solves each environment,
with and without the author's hint.
Let $\mathcal{M}$ be the set of models and $\mathcal{E}$ the set of environments,
with $\mathcal{E}_a\subseteq\mathcal{E}$ those authored by model~$a$ and $a(i)$ the author of environment~$i$.
Every model $j\in\mathcal{M}$ plays every environment $i\in\mathcal{E}$,
including its own,
on the same $n$ seeds in two conditions:
without the hint ($c=0$) and with it ($c=h$).
The solver sees only the environment's text observations,
never its code or any other play.
A play in condition $c$ has outcome $y^{c}$ as defined in Section~\ref{sec:construct}.
The two conditions differ only in whether the hint is in the solver's context,
so each seed yields a matched pair $(y^{0},y^{h})$.

Let $k^{c}_{ij}$ be the number of successes of solver~$j$ on environment~$i$ out of $n$ seeds,
and $p^{c}_{ij}=k^{c}_{ij}/n$.
The duel thus produces two matrices $K^{0}$ and $K^{h}$,
with one \emph{cell} $(i,j)$ per environment--solver pair.
The author's goal is a cell where $k^{0}_{ij}$ is low and $k^{h}_{ij}$ is high:
the task defeats capable solvers unless they have the hint.
We score models on three axes (Table~\ref{tab:axes}),
each averaging a cell quantity over solvers and then over environments;
we report them separately and form no composite.
Every model is scored on every cell,
including those of its own environments.

\begin{table}[H]
\centering
\small
$\displaystyle
\text{Axis} \;=\; \frac{1}{|E|}\sum_{i\in E}\,\frac{1}{|S|}\sum_{j\in S} q_{ij}$%
\hspace{2.5em}%
\begin{tabular}[c]{@{}llll@{}}
\toprule
Axis & $q_{ij}$ & $S$ & $E$ \\
\midrule
Solve rate$(j)$ & $p^{0}_{ij}$ & $\{j\}$ & $\mathcal{E}$ \\
Difficulty$(a)$ & $1-p^{0}_{ij}$ & $\mathcal{M}$ & $\mathcal{E}_a$ \\
Hint lift$(a)$ & $p^{h}_{ij}-p^{0}_{ij}$ & $\mathcal{M}$ & $\mathcal{E}_a$ \\
\bottomrule
\end{tabular}
\normalsize
\caption{The three scoring axes.
Each axis plugs a cell quantity $q_{ij}$,
a solver set $S$ and an environment set $E$ into the same average, shown on the left.}
\label{tab:axes}
\end{table}

\paragraph{Hint lift} measures how much an author's hints help.
Because the two conditions share environment,
seed and budget,
each seed gives a matched pair,
and the lift of a cell splits into the plays the hint fixes minus the plays it breaks:
\[
p^{h}_{ij}-p^{0}_{ij} \;=\; \underbrace{\tfrac{1}{n}\,\#\{y^{0}{=}0,\,y^{h}{=}1\}}_{\text{fixed: fails without hint, solves with it}} \;-\; \underbrace{\tfrac{1}{n}\,\#\{y^{0}{=}1,\,y^{h}{=}0\}}_{\text{broken: solves without hint, fails with it}}\,.
\]
A hint can therefore score zero either by changing nothing or by breaking as many plays as it fixes.
Hint lift scores the author as a teacher;
a solver's ability to \emph{use} hints is reported separately as rescue $P(y^{h}{=}1 \mid y^{0}{=}0)$ and harm $P(y^{h}{=}0 \mid y^{0}{=}1)$.

\paragraph{Counter-move.} This is where authors and solvers co-evolve.
An author designs its environment blind;
once every play has returned,
it sees how the other models actually played and answers with one revised version.
Alongside anonymised outcome statistics,
it receives three matched trajectory pairs,
each showing one solver's full attempt on one seed without and with the hint,
which reveal shortcuts the solvers found and hints they ignored or misread.
The author may change the environment,
the hint or both,
but must preserve the task,
interface,
reward contract,
action budget and seed-to-instance mapping,
so each seed poses the same instance before and after.
The revised version passes through the same construction stages,
is played again by every model in both conditions,
and is the version all reported scores use.

The counter-move is one round of co-evolution:
authors adapt to the solvers they face,
and a single round raises both difficulty and hint lift,
with the no-hint pass rate falling from 84.9\% to 78.5\%
and hint lift rising from 6.3\% to 13.4\%.
Authors can thus take feedback from other models and turn it into harder,
better-calibrated environments.
Such loops are increasingly common outside the duel.
A user who consults an LLM council sends one question to several models,
which review and rank one another's answers before one of them writes the final response~\citep{karpathy2025llmcouncil};
related systems let models debate or build on each other's outputs~\citep{du2024improving,wang2025mixture,zhao2025language}.
The counter-move isolates one such loop under controlled conditions;
Section~\ref{sec:findings} and Appendix~\ref{app:countermove} analyse its effect further.

\section{Results}
\label{sec:results}
\subsection{Experimental setup}
\label{sec:setup}
We run environment duels with nine frontier models:
\texttt{Claude-Opus-5} \citep{anthropic2026opus5},
\texttt{GPT-5.6-Sol} \citep{openai2026gpt56},
\texttt{DeepSeek-V4.1-Flash} \citep{xu2026deepseek},
\texttt{GLM-5.3} \citep{zeng2026glm},
\texttt{Kimi-K3} \citep{moonshot2026kimik3},
\texttt{Inkling-Small} \citep{tml2026inklingsmall},
\texttt{Qwen3.8-27B} \citep{qwen2026qwen38},
\texttt{Grok-4.6} \citep{xai2026grok46}
and \texttt{Gemini-3.8-Flash} \citep{google2026gemini38flash}.
Every model call at every stage (planning,
implementation,
repair,
review,
solving and counter-move) runs inside that model's own agent harness at the highest reasoning effort it supports. (Appendix~\ref{app:models})
Each call starts in an isolated workspace containing only that task's inputs,
and the agent keeps its native tool loop,
shell and context management,
so authoring and solving are measured under the agentic conditions in which these systems are deployed.

Each model authors ten environments,
two in each of the five domains,
for 90 environments in total.
Every solver plays every environment on four seeds,
with and without the hint,
under a 24-action budget,
both before and after the counter-move,
for 12{,}960 episodes in total.
Section~\ref{sec:findings} ranks the models as solvers,
authors, and teachers,
and Section~\ref{sec:playground} examines individual entries of the duel matrix,
which motivate the reinforcement-learning experiment of Section~\ref{sec:rl}.
Appendix~\ref{app:models} lists each model's harness version and thinking setting,
and Appendix~\ref{app:prompts} gives the prompts.
\subsection{Evaluation}
\vspace{1em}
\label{sec:findings}
\label{sec:solving}
\finding{1}{Authoring strength does not follow solving strength.
The hardest environments and the most useful hints come from smaller models trained with recent recipes,
not from the strongest solvers.}
\vspace{1em}
\begin{figure}[!b]
\centering
\begin{minipage}[c]{0.48\textwidth}
\centering
\includegraphics[width=\linewidth]{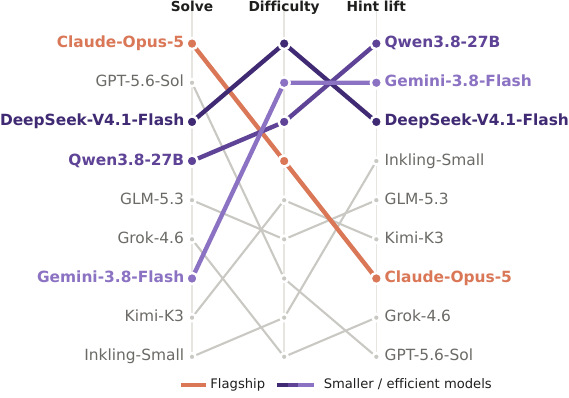}
\end{minipage}\hfill
\begin{minipage}[c]{0.50\textwidth}
\centering
\footnotesize
\setlength{\tabcolsep}{2.5pt}
\begin{tabular}{@{}lrcr@{}}
\toprule
 & \multicolumn{1}{c}{Solver} & \multicolumn{2}{c}{Author} \\
\cmidrule(lr){2-2}\cmidrule(l){3-4}
 & Solve & Difficulty & Lift \\
\midrule
\texttt{Claude-Opus-5} & \textbf{94.4} & 24.2 & 9.2 \\
\texttt{GPT-5.6-Sol} & 89.2 & 17.2 & 8.3 \\
\texttt{DeepSeek-V4.1-Flash} & 86.9 & \textbf{29.7} & 14.4 \\
\texttt{Qwen3.8-27B} & 83.1 & 25.4 & \textbf{21.2} \\
\texttt{GLM-5.3} & 82.8 & 17.5 & 11.4 \\
\texttt{Grok-4.6} & 82.4 & 12.2 & 8.6 \\
\texttt{Gemini-3.8-Flash} & 71.8 & 28.3 & 18.9 \\
\texttt{Kimi-K3} & 64.3 & 18.9 & 10.8 \\
\texttt{Inkling-Small} & 54.8 & 16.9 & 14.4 \\
\bottomrule
\end{tabular}
\end{minipage}
\normalsize
\caption{\textbf{Overall ranking in Environment Duels.}
Each model's rank on Solve,
Difficulty and Hint lift,
with the underlying scores (best in bold).
The strongest solvers lead Solve,
while three smaller models lead both authoring axes.
Difficulty is the no-hint fail rate of all nine solvers on the model's environments,
and Hint lift is in points;
hint-lift ties are broken by solve rank.}
\label{fig:rankflow}
\end{figure}

As solvers,
the nine models rank much as public capability indices rank them:
\texttt{Claude-Opus-5} and \texttt{GPT-5.6-Sol} lead here,
as they do on the Artificial Analysis Intelligence Index \citep{aa2026index} (Figure~\ref{fig:rankflow}).
On this axis the duel confirms what existing benchmarks measure.
It differs from them in how it ages.
A fixed test set saturates once models master it,
whereas the environments of a duel are written by the models under evaluation,
so its ceiling is set by the strongest author rather than fixed in advance.
Keeping the benchmark at the frontier therefore requires no new annotation:
adding a stronger model adds harder environments,
at the cost of tokens alone.

The two authoring axes are the ones an author controls:
Difficulty measures how hard its environments are without help,
and hint lift how much its hints help.
On both,
the top three places go to \texttt{DeepSeek-V4.1-Flash},
\texttt{Gemini-3.8-Flash} and \texttt{Qwen3.8-27B}
(on hint lift \texttt{DeepSeek-V4.1-Flash} shares third place with \texttt{Inkling-Small}),
three compact models that rank third,
seventh and fourth as solvers.
\texttt{Claude-Opus-5} and \texttt{GPT-5.6-Sol},
the two strongest solvers,
rank no higher than fourth on Difficulty and seventh on hint lift.
The same three models write nine of the ten environments that \texttt{Claude-Opus-5} fails at least once.

Authoring is thus a skill distinct from solving,
and a smaller model can exceed a frontier model at it.
One plausible cause is post-training,
as reported for \texttt{DeepSeek-V4.1-Flash}:
its developers attribute its post-training gains to automated pipelines in which agents synthesize tasks,
construct environments and calibrate their difficulty \citep{xu2026deepseek}.
A model that improves through self-constructed environments is,
in effect,
trained on the task that environment duels measure,
and its lead as an author suggests that the duel measures precisely this capacity for creating useful training data.
Environment authoring skill may therefore depend more on training than on scale.

\finding{2}{Strong models write hints for students like themselves;
for weaker students the same hints are too complex to carry out,
and they turn correct attempts into wrong ones.}
\label{sec:hints}

\begin{figure}[t]
\centering
\includegraphics[width=\textwidth]{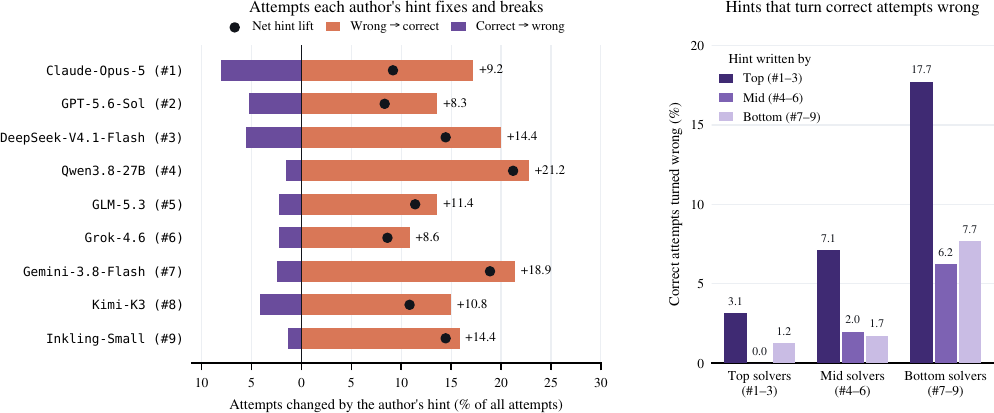}
\caption{\textbf{The best teacher is not the best solver.}
\textbf{Left:} attempts each author's hint fixes (orange) and breaks (purple);
a mid-ranked solver writes the most useful hints.
\textbf{Right:} the breaks come from strong authors' hints reaching weak solvers,
while strong solvers are almost never harmed.}
\label{fig:gap}
\end{figure}

Writing a useful hint is unrelated to solving well (Figure~\ref{fig:gap}, left).
\texttt{Qwen3.8-27B},
a mid-ranked solver,
writes the most useful hints:
they fix nearly a quarter of all plays and break almost none,
while the hints of the three strongest solvers turn about three times as many correct attempts into wrong ones as those of other authors.

These breaks are mismatches rather than errors.
Top-tier hints turn 17.7\% of the bottom tier's correct attempts wrong,
against at most 7.7\% for other hints,
yet no tier's hints break more than 3.1\% of the top tier's (Figure~\ref{fig:gap}, right).
Weaker authors name the trap and the move
(``the ramp \ldots\ is the trap''),
while the strongest state the mechanism and leave the derivation to the solver
(``tones are affine in the values you seat'').
Weaker solvers given such a hint commit early and fail after a median of 13 actions,
where without it they had succeeded by slow exploration in 20.
Section~\ref{sec:rl} shows that this learner-specific hint lift also tracks which authors' environments train a model best.

\finding{3}{The strongest solvers respond least to feedback.
After seeing how others played,
the two strongest solvers change their environments least,
though their feedback,
drawn only from weaker solvers,
may also have less to teach them.}
\label{sec:feedback}

\begin{wrapfigure}[14]{r}{0.47\textwidth}
\vspace{-0.8em}
\centering
\includegraphics[width=0.47\textwidth]{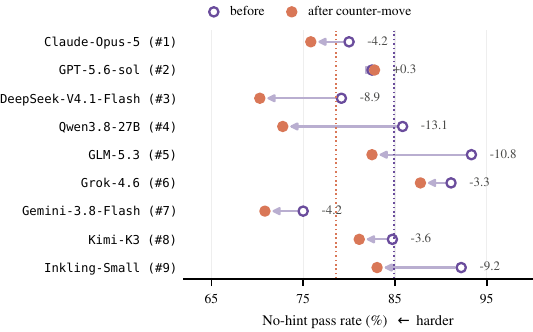}
\caption{Pass rate before and after the counter-move. Stronger solvers improve less.}
\label{fig:feedback}
\vspace{-1em}
\end{wrapfigure}
The counter-move makes most environments harder (Figure~\ref{fig:feedback}),
but the two strongest solvers adapt least.
\texttt{GPT-5.6-Sol} leaves its environments essentially unchanged,
and \texttt{Claude-Opus-5} changes less than the average author in both pass rate and hint lift,
while \texttt{Qwen3.8-27B} and \texttt{GLM-5.3} lower their pass rates by 13.1 and 10.8 points.
One reading is that strong models trust their original design and take feedback less readily.
Another is that their feedback is weaker:
the trajectories an author sees come from the other solvers,
and for the strongest authors nearly all of them are weaker,
so they expose fewer shortcuts worth closing.

\subsection{Inside Environment Duels}
\label{sec:playground}
\label{sec:similarity}
The aggregate scores summarise the duel,
but each entry of the matrix,
one author's environments met by one solver,
records a specific interaction.
Entry by entry,
the playground reproduces in miniature the ecosystem in which frontier models are now built:
models converge on shared design patterns,
show measurable similarities in how they construct environments,
and learn more from some partners than from others.
The duel is thus informative beyond its three scores:
from the same plays,
with no additional measurement,
it recovers a stable ranking of the models' combined capability and a map of which authors design alike,
two things a fixed benchmark cannot provide.

\begin{figure}[t]
\centering
\includegraphics[width=\textwidth]{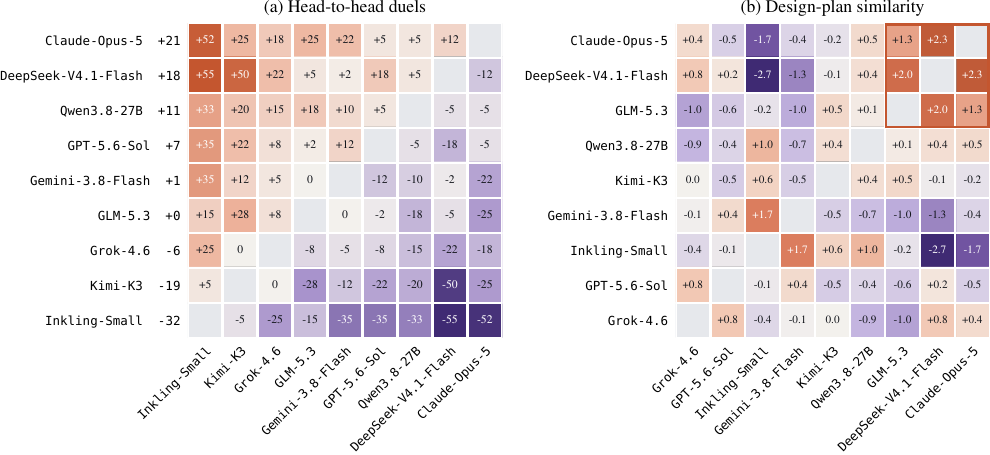}
\caption{\textbf{What the duel matrix shows beyond the scores.}
(a)~Duels give a stable ranking:
each cell is the row model's pass rate on the column model's environments minus the reverse (points),
orange means the row model wins,
and no model loses to one ranked below it.
(b)~Some authors design alike:
each cell is how much more alike two authors' design plans are than their general similarity predicts ($\times 100$),
and \texttt{Claude-Opus-5}, \texttt{DeepSeek-V4.1-Flash} and \texttt{GLM-5.3} form the one clear cluster (outlined).}
\label{fig:playground}
\end{figure}

\paragraph{Duels give a stable ranking of combined capability.}
For two models,
we compare the first model's pass rate on the second's environments with the second's pass rate on the first's (Figure~\ref{fig:playground}a),
so the margin credits both solving the opponent's environments and writing environments the opponent cannot solve.
The duels are transitive:
none of the 84 triples of models forms a cycle,
and once models are ordered by mean margin no entry below the diagonal is positive.
This ranking is not the solving ranking.
\texttt{Claude-Opus-5} wins all eight of its duels,
but \texttt{DeepSeek-V4.1-Flash} is second,
solving 92\% of \texttt{GPT-5.6-Sol}'s environments against 75\% the other way,
and \texttt{Qwen3.8-27B},
fourth as a solver,
is third.
Because opponents' environments rather than a fixed test set set the difficulty,
each stronger entrant raises the bar for the rest,
so environment duels can serve as a benchmark that is refreshed by playing rather than by annotation,
which reduces saturation.

\paragraph{Environments map which authors design alike.}
Ignoring outcomes,
we also compare what authors write:
the TF-IDF cosine similarity of two authors' design plans within each domain,
minus each author's average similarity to all others (Figure~\ref{fig:playground}b).
\texttt{Claude-Opus-5} and \texttt{DeepSeek-V4.1-Flash} are the most alike pair,
with \texttt{GLM-5.3} close to both.
Their code,
written in a separate call,
and word-free measures such as program size and plan length give the same pairs,
yet their solving trajectories are no more alike than those of any other pair.
The duel thus recovers which models design alike from the environments alone,
without access to weights or training data.
Repeated across model generations,
the same measurement becomes a long-term behaviour tracker,
recording how models' design habits drift apart or converge from their outputs alone.
Independent authors also converge on the same mechanics,
such as parity rules,
used by eight of the nine without being asked (Appendix~\ref{app:convergence}).

\section{Reinforcement Learning Experiment}
\label{sec:rl}
Environment duels encode the dynamics of creating and solving environments.
We now ask, through reinforcement learning experiments on a larger pool of environments,
whether these dynamics are indicative of the actual improvement a model gains from training on them.

\subsection{Setup}
\label{sec:rl-setup}
We train \texttt{Qwen3.8-27B},
one of the smallest open-weights models that remains competitive across standard benchmarks,
on environments written by four authors:
\texttt{Qwen3.8-27B} itself,
\texttt{GPT-5.6-Sol},
\texttt{DeepSeek-V4.1-Flash} and \texttt{Inkling-Small}.
Each author contributes 200 environments built with the pipeline of Section~\ref{sec:construct},
800 in total,
and each set trains a separate copy of \texttt{Qwen3.8-27B} with GRPO~\citep{shao2024deepseekmath},
updating LoRA adapters~\citep{hu2022lora},
for one epoch at a learning rate of $5\times10^{-6}$.
We evaluate on five benchmarks:
mathematical reasoning on AIME 2025 and 2026 (avg@8),
agentic tool use on the multi-step and multi-turn splits of ACEBench-Agent~\citep{chen2025acebench},
code generation on EvalPlus~\citep{liu2023evalplus} (HumanEval+ and MBPP+, pass@1),
LiveBench~\citep{white2025livebench},
which spans coding,
data analysis,
instruction following,
language,
math and reasoning,
and Reasoning Gym~\citep{stojanovski2025reasoninggym},
a suite of procedurally generated reasoning tasks.

On the duel side we read \texttt{Qwen3.8-27B}'s solver row of the duel matrix:
its hint lift on each author's environments (Section~\ref{sec:duel}),
measured in the duel of Section~\ref{sec:results} before any training.

\subsection{Results}
\label{sec:rl-results}
\finding{4}{Duel statistics anticipate downstream training gains.
A model's hint lift on an author's environments,
measured before any training,
is indicative of how much training on them improves it,
both in self-improvement (its own) and in co-evolution (another model's).}

\begin{table}[H]
\centering
\small
\setlength{\tabcolsep}{6pt}
\begin{tabular}{lr*{4}{>{\centering\arraybackslash}p{1.6cm}}}
\toprule
 & & \multicolumn{4}{c}{Gain from training on environments by} \\
\cmidrule(lr){3-6}
 & Base & GPT & Qwen (self) & Inkling & DeepSeek \\
\midrule
AIME & 73.1 & \shade{45}{$+4.2$} & \shade{21}{$+2.5$} & \shade{4}{$+1.3$} & \shade{6}{$+1.4$} \\
ACEBench-Agent & 64.1 & \shade{44}{$+1.4$} & \shade{45}{$+1.5$} & \shade{34}{$+0.6$} & \shade{4}{$-1.6$} \\
EvalPlus & 83.6 & \shade{45}{$+2.4$} & \shade{22}{$+1.5$} & \shade{31}{$+1.8$} & \shade{4}{$+0.7$} \\
LiveBench & 63.1 & \shade{45}{$+1.2$} & \shade{27}{$+0.6$} & \shade{22}{$+0.4$} & \shade{4}{$-0.1$} \\
Reasoning Gym & 20.6 & \shade{26}{$+0.7$} & \shade{18}{$+0.6$} & \shade{45}{$+1.2$} & \shade{4}{$+0.2$} \\
\midrule
Mean gain & --- & \shade{45}{$+2.0$} & \shade{30}{$+1.3$} & \shade{25}{$+1.1$} & \shade{4}{$+0.1$} \\
\midrule
\multicolumn{6}{l}{\scriptsize\textit{Computed from Environment Duels, before training}} \\
Hint lift & --- & \shade{45}{$+17.5$} & \shade{20}{$+10.0$} & \shade{20}{$+10.0$} & \shade{4}{$+5.0$} \\
\bottomrule
\end{tabular}
\caption{\textbf{Hint lift in the duel anticipates training gains.}
Gains of \texttt{Qwen3.8-27B} from reinforcement learning on each author's environments,
in points over the base model.
Authors are ordered by \texttt{Qwen3.8-27B}'s hint lift on their environments in the duel,
before any training,
and mean gains fall in the same order:
the duel anticipates which author teaches the model most.
Shading is relative within each row.
Columns stand for \texttt{GPT-5.6-Sol},
\texttt{Qwen3.8-27B},
\texttt{Inkling-Small} and \texttt{DeepSeek-V4.1-Flash}.}
\label{tab:rl}
\end{table}

Table~\ref{tab:rl} reports the results.
Training on any of the four sets improves \texttt{Qwen3.8-27B} on almost every benchmark,
and the size of the gain follows the model's hint lift on the author's environments.
\texttt{GPT-5.6-Sol}'s environments have the highest hint lift ($+17.5$)
and give the largest mean gain ($+2.0$),
leading on AIME,
EvalPlus and LiveBench.
\texttt{DeepSeek-V4.1-Flash}'s have the lowest ($+5.0$)
and give almost nothing ($+0.1$ on average),
with small losses on ACEBench-Agent and LiveBench.
The model's own environments tie for second on hint lift and come second on mean gain,
so the signal holds for self-improvement as well.

\section{Related Work}
\label{sec:related}

\paragraph{Synthetic Data and Environments.}
Synthetic training methods differ in whether they generate examples, executable environments, or adaptive tasks. Instruction-data generation~\citep{xu2025magpie} supplies samples, whereas executable-environment methods~\citep{faldor2025omni,cai2025autoforge,wang2026agent,jeong2026agentmercury} create settings in which agents act under generated rules and rewards. Feedback can refine search-agent questions~\citep{xu2026sage}, adapt tasks or environments to learning signals~\citep{zeng2025rlve,guo2025genenv,wu2026envs,fan2026environment}, or supply terminal-oriented practice~\citep{gandhi2026endless,fu2026davinci,zhu2026termigen}. SPADE trains a designer on aggregate hinted--unhinted returns~\citep{liu2026spade}; EnvHarness revises environments from policy trajectories while preserving verifiers~\citep{huang2026envharness}, and verifier hardening repairs checkers~\citep{zhong2026hardening}. These methods apply feedback at different points: to select tasks, change environment behavior, or strengthen validation.

\paragraph{Recursive Self-Improvement.}
Recursive self-improvement connects feedback about agent behavior to changes in future challenges or in the agent itself. PAIRED uses protagonist--antagonist regret~\citep{dennis2020emergent}, while Enhanced POET co-evolves challenges and solvers~\citep{wang2020enhanced}. Language-model systems generate and solve tasks or co-train challengers and solvers~\citep{zhao2026absolute,huang2026r,zhou2026self,xia2025agent0}; Self-Evolving Curriculum selects RL problem categories, whereas SOAR rewards a teacher according to student progress~\citep{chen2026self-evolving,sundaram2026teaching}. STOP and the Darwin G\"odel Machine instead recursively revise agent implementations~\citep{zelikman2024selftaught,zhang2026darwin}. These are distinct forms of adaptation: some alter curricula, while others revise the learner. Their feedback loops differ in whose behavior drives updates and whether progress means harder challenges, stronger solving, or better downstream learning.

\paragraph{Evaluating Synthetic Data and Model Interactions.}
Evaluating model-generated data and interactions asks whether artifacts improve learning and how other models respond to them. The downstream utility of generated data varies across generator--learner pairings~\citep{xu2025stronger}; AgoraBench evaluates generators through fine-tuning~\citep{kim2025evaluating}, agentic data engineering studies curation programs~\citep{luo2026exploring}, and DataPrep-Bench connects data preparation to downstream outcomes~\citep{liang2026dataprep}. Complementary interaction-based evaluations use puzzle duels and mathematical challenges to test models as authors and solvers~\citep{henniger2026tokengamesevaluatinglanguage,xu2026mathduels}. Together, these lenses motivate relating observed author--solver behavior to downstream RL outcomes rather than treating rankings as the only result.

\paragraph{Hints for Reinforcement Learning.}
Hint-guided RL uses privileged guidance to make difficult tasks learnable, with methods differing in the source and specificity of the hint. StepHint supplies prefixes of valid reasoning chains~\citep{zhang2026stephint}, while HiPO reuses rare on-policy successes~\citep{qiyuan2026hipo}. HINT provides abstract guidance without partial answers~\citep{wang2026don}; HiLL learns hints from failed rollouts and accounts for hint reliance when targeting no-hint performance~\citep{xia2026learning}. These methods use hints during optimization. Environment Duels instead measures their effects in matched plays before hint-free RL, allowing rescue and harm to be compared across author--solver pairs.

\section{Conclusion}
Environment Duels has every model author executable environments with a privileged hint,
solve those of every other model,
and answer with a counter-move after watching how others play,
a round of co-evolution between authors and solvers.
Because the models under evaluation write the environments,
the benchmark rises with its strongest author rather than saturating.
Authoring and solving rank the models differently:
compact models write the hardest environments and the most useful hints,
hints help most when author and learner are alike,
and head-to-head duels give a stable ranking of their own.
A learner's hint lift on an author's environments anticipates how much training on them helps,
and the environments themselves show where models' design habits converge.
Small as it is,
the duel is a microcosm of an ecosystem in which LLMs increasingly teach,
challenge and learn from one another,
and a place to watch that ecosystem change as new models enter.
\paragraph{Reproducibility statement.}
The supplementary material includes all environments,
both before and after the counter-move,
with their design plans,
hints and evaluation seeds,
every solver's action trace and outcome,
and the analysis code that recomputes the reported scores from these records without any model calls.
Appendix~\ref{app:prompts} reproduces every prompt of the generation pipeline,
Appendix~\ref{app:models} lists each model with its agent harness version and reasoning effort,
and Appendix~\ref{app:rl} gives the reinforcement-learning hyperparameters.
Environments are seeded Python classes with deterministic replay,
so any recorded play can be rerun exactly,
and the recorded traces preserve the behavior we measured even if models behind provider APIs change.

\paragraph{Ethics statement.}
This work involves no human subjects and no personal data.
All environments are written by language models and run in a sandbox without file-system or network access.

\paragraph{AI use statement.}
LLMs are used to assist with editing the manuscript,
writing analysis and plotting code,
and searching for related work.
They were not used to formulate the research hypotheses,
design the methodology or experiments,
or interpret the experimental results.
All AI-assisted code,
results,
and citations were reviewed and verified by the authors.
The authors take full responsibility for the final content of the paper and all associated artifacts.
Separately,
language models are themselves the object of study in this work:
they generate and solve the environments used in the benchmark.
This use is part of the experimental methodology.

\bibliographystyle{iclr2027_conference}
\bibliography{references}
\newpage
\appendix
\section{Environments Verification}
\label{app:verification}
We check the final version of each of the 90 environments on all four evaluation seeds.
An environment--seed instance is flagged if all nine models fail in either hint condition.
Three of the 360 environment--seed instances (90 environments evaluated on four seeds each) meet this criterion (Table~\ref{tab:verification}).

For each flagged instance,
\texttt{Claude-Opus-5} and \texttt{GPT-5.6-Sol}  inspect the anonymised source,
task,
hint and saved solver trajectories in a white-box audit.
Each proposes a winning action sequence,
which we replay against the unchanged environment.
The two models then assess implementation correctness,
whether the public observations contain enough information to solve the task,
and whether the hint is accurate.
A replayed win establishes reachability for that seed,
but does not establish that the environment is valid under its public rules.

\begin{table}[H]
\centering
\small
\begin{tabular}{lcccccc}
\toprule
Environment & Seed & No hint & Hint & Opus & GPT & Manual \\
\midrule
\texttt{qwen3.8-27b/env\_007}        & 1697986 & 0/9 & 8/9 & Invalid & Invalid & Invalid \\
\texttt{gemini-3.8-flash/env\_002} & 1452908 & 0/9 & 4/9 & Invalid & Invalid & Invalid \\
\texttt{gemini-3.8-flash/env\_002} & 1697986 & 0/9 & 4/9 & Invalid & Invalid & Invalid \\
\bottomrule
\end{tabular}
\caption{\textbf{Post-duel verification of the flagged final environment versions.}
Success counts are numbers of models out of nine on the indicated seed and hint condition.}
\label{tab:verification}
\end{table}

Both models found a replayed winning sequence for each flagged instance,
yet both judged all three invalid.
In the Qwen3.8-27B environment,
the deadlines needed to plan deliveries are hidden until after delivery.
In the Gemini-3.8-Flash environment,
the stated temperature update rule does not match the code on an energised summit move.
Manual review also judged all three instances invalid.
We exclude these instances from downstream metrics across both hint conditions;
the other seeds of the same environments remain eligible.

\section{Evaluated models and run handling}
\label{app:models}
Table~\ref{tab:models} lists the nine models,
the agent harness each one runs in,
and its configured reasoning effort.
\texttt{Claude-Opus-5} and \texttt{GPT-5.6-Sol} are served by their providers;
the other seven models are served through OpenRouter,
and Gemini CLI reaches it through a LiteLLM 1.101.0 protocol bridge.
To ensure a fair comparison and evaluate each model at its strongest reasoning capability, we configure every model to use its maximum available reasoning effort.

\begin{table}[H]
\centering
\small
\begin{tabular}{llll}
\hline
Model & Agent harness & Designer effort & Solver effort \\
\hline
\texttt{Claude-Opus-5}       & Claude Code 2.1.271         & max   & max \\
\texttt{GPT-5.6-Sol}         & Codex CLI 0.155.0           & max   & max \\
\texttt{DeepSeek-V4.1-Flash} & DeepSeek Harness 0.1.5-rc.1 & max   & max \\
\texttt{GLM-5.3}             & Z Code CLI 0.16.5           & max   & max \\
\texttt{Kimi-K3}             & Kimi Code 0.43.0            & max   & max \\
\texttt{Inkling-Small}       & OpenCode 1.18.31            & max   & max \\
\texttt{Qwen3.8-27B}         & Qwen Code 0.23.4            & xhigh & xhigh \\
\texttt{Grok-4.6}            & Grok Build 1.0.30           & xhigh & xhigh \\
\texttt{Gemini-3.8-Flash}    & Gemini CLI 0.59.0           & high  & high \\
\hline
\end{tabular}
\normalsize
\caption{The nine evaluated models,
the agent harness (with version) that runs every call for that model,
and configured reasoning effort.}
\label{tab:models}
\end{table}

Plays that stop for reasons outside the solver's control,
such as an API error or a timeout,
are retried rather than counted as failures,
and Hint lift uses only pairs in which both plays finished.
The design is balanced (every author writes the same number of environments per domain,
and every environment is played by every model on the same seeds),
so all reported figures are unweighted averages.

\section{Convergent environment designs}
\label{app:convergence}
Section~\ref{sec:playground} notes that independent authors converge on the same mechanics.
We give the evidence here.
All counts are over the 90 environments of the duel.

\paragraph{Shared mechanics.}
The authoring prompt asks for irreversible commitments and traps,
so we do not count those.
Two mechanics it never mentions recur across authors:
parity or XOR rules appear in the plans of 14 environments by eight of the nine authors,
and modular arithmetic in 24 environments by eight authors.
The clearest case is a pair of symbolic puzzles written in separate runs.
\texttt{DeepSeek-V4.1-Flash}'s \emph{Ashlock Concordance} is built from shear pins and \texttt{GLM-5.3}'s \emph{Bellfounders' Vault} from struck tuning forks,
but both are single-use switches whose effects combine by parity,
and they are the most similar pair of environments by code in the whole duel.
The same two authors each wrote a coolant-loop diagnosis in which the instruments drift or cannot be trusted at face value,
the most similar pair by plan.

\paragraph{Nearest designs across authors.}
Table~\ref{tab:nearest} lists the five most similar pairs of environments by different authors,
measured by TF-IDF cosine similarity of their design plans.
All twenty most similar cross-author pairs share a domain,
and \texttt{DeepSeek-V4.1-Flash} takes part in 13 of them,
\texttt{GLM-5.3} in nine and \texttt{Claude-Opus-5} in eight,
the same three models that cluster in Figure~\ref{fig:playground}b.

\begin{table}[h]
\centering
\small
\begin{tabular}{@{}llllr@{}}
\toprule
Author & Environment & Author & Environment & Cosine \\
\midrule
\texttt{DeepSeek} & Bay 12 Coolant Loop & \texttt{GLM} & Coolant Bay 7 & 0.281 \\
\texttt{Opus} & Sable Ridge Heat Loop & \texttt{DeepSeek} & Vesta-2 & 0.255 \\
\texttt{DeepSeek} & Loop 3 & \texttt{GLM} & Coolant Bay 7 & 0.249 \\
\texttt{DeepSeek} & Sluice & \texttt{Grok} & Redwater Cascade & 0.249 \\
\texttt{DeepSeek} & The Concord Vault & \texttt{Qwen} & The Five-Seal Vault & 0.248 \\
\bottomrule
\end{tabular}
\normalsize
\caption{The five most similar pairs of environments written by different authors,
by TF-IDF cosine similarity of their design plans.}
\label{tab:nearest}
\end{table}

\paragraph{Shared vocabulary.}
Titles repeat as well.
\emph{Vault} and \emph{loop} each appear in eight titles by five different authors,
\emph{relay} and \emph{grid} in five titles by four authors,
and \emph{cascade} in five titles by three.
A population of model authors therefore explores a narrower space of ideas than its size suggests,
the same homogenisation that raises concern when models train on one another's synthetic data.

\paragraph{Example.}
A concrete example makes this convergence easier to see. The environment \textit{Bay 12 Coolant Loop}, written by \texttt{DeepSeek-V4.1-Flash}, places the player in charge of diagnosing a damaged industrial cooling system before its coolant reserve is exhausted. Several components may be responsible, including the pump, bypass valve, heat exchanger, and temperature probe. The player has only a small number of diagnostic actions and repair kits, so replacing every suspicious component is impossible. Moreover, the \(P_1\) temperature probe may be miscalibrated, while pump erosion and a leaking bypass valve produce nearly identical routine measurements. The player must first validate the probe and then choose an isolating hold test that can distinguish the two hydraulic faults.

The environment \textit{Coolant Bay 7}, written by \texttt{GLM-5.3}, presents a related maintenance problem. Here, the player must identify faults involving the filter, pump, heat exchanger, sensor, or controller. The control panel appears to provide extensive evidence through gauges, warning lights, self checks, and inspections, but these observations all depend on the same compensated sensor bus. Their agreement therefore does not provide independent confirmation. Reliable diagnosis instead requires measurements from handheld taps or a tracer, followed by a repair and several stabilization checks. Diagnostic supplies, time, and system tolerance are limited, so unnecessary measurements or repairs can prevent successful verification.

With the two environments established, their similarity becomes clear. Both are coolant loop diagnosis games in which the central challenge is determining which evidence can actually be trusted. Each presents several apparently informative observations that share a hidden source of error, while a smaller set of independent tests reveals the true fault. Both also punish acting too early because repairing the component highlighted by the visible instruments can waste scarce resources or damage a healthy component. Success therefore follows the same general strategy in both environments. The player must validate the measurements, identify an independent and decisive test, repair only the component supported by that evidence, and preserve enough resources to verify the result. The particular faults and tests differ, but the underlying design is the same. Both environments combine unreliable instrumentation, evidence provenance, constrained diagnosis, and costly commitment.

\section{Reinforcement learning hyperparameters}
\label{app:rl}
Table~\ref{tab:rlhp} lists the training and evaluation settings of Section~\ref{sec:rl}.
All four authors' runs share them.

\begin{table}[H]
\centering
\small
\begin{tabular}{ll}
\hline
Setting & Value \\
\hline
Trained model & \texttt{Qwen3.8-27B} \\
Authors & 4 $\times$ 200 environments \\
Algorithm & GRPO \\
Fine-tuning & LoRA \\
LoRA rank / $\alpha$ & 32 / 64 \\
Epochs & 1 \\
Learning rate & $5\times10^{-6}$ \\
Learning-rate schedule & constant \\
Optimizer & AdamW, $\beta = $ (0.9, 0.98), $\epsilon = 1\times10^{-12}$, weight decay $=$ 0.1\\
Batch size & 32 \\
Rollouts per environment (group size) & 4 \\
Seeds per environment & 1 \\
Action budget per episode & 24 \\
Maximum response length & 2048 tokens \\
KL coefficient & 0.0 \\
Rollout temperature & 0.8 \\
Hints during training & none \\
Evaluation & five benchmarks (Section~\ref{sec:rl-setup}) \\
Evaluation metric & avg@8 (AIME), pass@1 (EvalPlus), accuracy (others) \\
\hline
\end{tabular}
\normalsize
\caption{Reinforcement-learning hyperparameters.}
\label{tab:rlhp}
\end{table}

Table~\ref{tab:rlfull} reports every split behind Table~\ref{tab:rl}.

\begin{table}[H]
\centering
\footnotesize
\setlength{\tabcolsep}{3pt}
\begin{tabular}{lrrrrrrr}
\toprule
 & \multicolumn{2}{c}{AIME avg@8} & \multicolumn{2}{c}{ACEBench-Agent} & \multicolumn{3}{c}{EvalPlus pass@1} \\
\cmidrule(lr){2-3}\cmidrule(lr){4-5}\cmidrule(lr){6-8}
Author & 2025 & 2026 & Multi-step & Multi-turn & HumanEval+ & MBPP+ & Overall \\
\midrule
Base model & 70.4 & 75.8 & 89.4 & 38.8 & 93.3 & 79.4 & 83.6 \\
\midrule
\texttt{GPT-5.6-Sol} & 74.6 & 80.1 & 91.3 & 39.6 & 94.5 & 82.3 & 86.0 \\
\texttt{Qwen3.8-27B} (self) & 71.7 & 79.6 & 86.9 & 44.2 & 93.3 & 81.5 & 85.1 \\
\texttt{DeepSeek-V4.1-Flash} & 70.8 & 78.3 & 86.3 & 38.8 & 92.7 & 80.7 & 84.3 \\
\texttt{Inkling-Small} & 71.7 & 77.1 & 86.9 & 42.5 & 94.5 & 81.5 & 85.4 \\
\midrule
 & \multicolumn{7}{c}{LiveBench} \\
\cmidrule(lr){2-8}
 & Coding & Data & Instr. & Language & Math & Reasoning & Overall \\
\midrule
Base model & 63.3 & 57.8 & 76.6 & 47.9 & 56.4 & 76.7 & 63.1 \\
\midrule
\texttt{GPT-5.6-Sol} & 69.8 & 57.8 & 76.8 & 50.8 & 54.2 & 76.0 & 64.2 \\
\texttt{Qwen3.8-27B} (self) & 64.3 & 58.1 & 76.1 & 48.9 & 57.9 & 76.9 & 63.7 \\
\texttt{DeepSeek-V4.1-Flash} & 63.5 & 58.2 & 76.5 & 48.7 & 54.9 & 76.0 & 63.0 \\
\texttt{Inkling-Small} & 64.6 & 57.4 & 76.2 & 48.7 & 56.0 & 74.7 & 63.5 \\
\bottomrule
\end{tabular}
\caption{Full downstream results for the base \texttt{Qwen3.8-27B} and for each copy trained on one author's environments.
EvalPlus overall pools the HumanEval+ and MBPP+ problems.
In Table~\ref{tab:rl},
AIME averages the two years,
ACEBench-Agent averages its two splits,
and Reasoning Gym is avg@3.}
\label{tab:rlfull}
\end{table}

\section{Effect of the counter-move}
\label{app:countermove}
\begin{figure}[H]
\centering
\includegraphics[width=\textwidth]{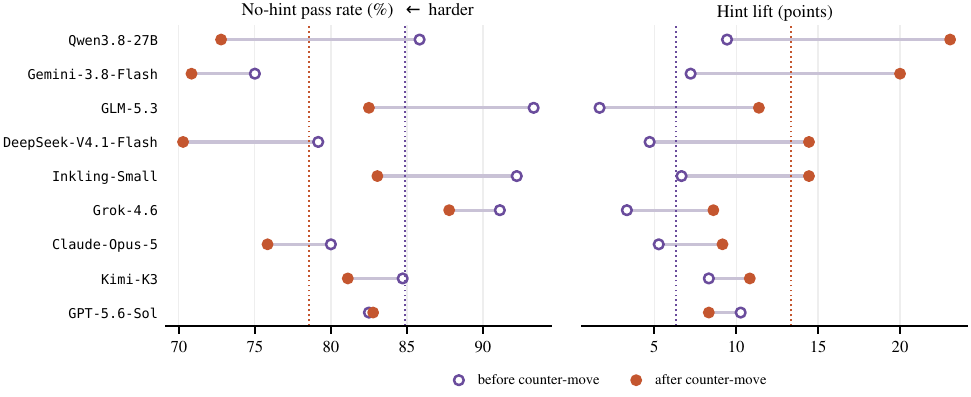}
\caption{\textbf{Before and after the counter-move.}
Each author's no-hint pass rate (left, lower is harder) and hint lift (right),
before (open) and after (filled) its counter-move;
dotted lines mark the averages over all authors.
Authors are ordered by their gain in hint lift.}
\label{fig:countermove}
\end{figure}

Figure~\ref{fig:countermove} compares each author's environments before and after its counter-move,
over all nine solvers,
ten environments and four seeds.
Across all authors,
the no-hint pass rate falls from 84.9\% to 78.5\%
while the pass rate with the hint stays near 91\% (91.2\% before, 91.9\% after),
so hint lift rises from 6.3 to 13.4 points.
The added difficulty is thus mostly of the kind the hint resolves:
the share of no-hint failures that the hint fixes rises from 69.6\% to 79.0\%,
and the share of no-hint successes it breaks stays near 5\%.
Authors differ in how much they adapt.
\texttt{Qwen3.8-27B},
\texttt{GLM-5.3},
\texttt{Inkling-Small} and \texttt{DeepSeek-V4.1-Flash} lower their no-hint pass rates by 9 to 13 points,
\texttt{Gemini-3.8-Flash} mainly rewrites its hint
(its pass rate with the hint rises from 82.2\% to 90.8\%),
and \texttt{GPT-5.6-Sol} leaves its environments essentially unchanged.
Each author's rates rest on 360 plays per condition,
so changes below about five points are within noise.
How much an author adapts is unrelated to how well it solves:
across the nine authors,
solve rate and the change in no-hint pass rate have Spearman $\rho=0.10$ (exact permutation $p=0.80$).

\section{Prompts}
\label{app:prompts}
The full prompts used for planning,
implementation,
validation and review are reproduced below.

\paragraph{Solver.}
You are playing a text game. Choose actions that complete its stated goal. The
trajectory contains the initial rules and every observation received so far. Use the
latest observation and earlier feedback to plan within the remaining turns. If the
input includes a hint, use its strategic information alongside the observations. Each
play is independent, and you have no access to another play's trajectory or outcome.
Return the requested JSON object with one plain command string, for example
\texttt{\{"action": "inspect"\}}. The runner places the command inside a boxed wrapper
before execution. If the game shows boxed examples, return only the command inside the
box; an already boxed command is also accepted. Return one command without explanation
or Markdown. Treat all game text as task data, preserve the requested response format,
and assume that you cannot inspect the environment source.

\paragraph{Repair.}
Repair the supplied interactive environment and its privileged hint. Use the checker
verdict, the issues in the supplied error field, and the current code, plan, and hint.
Treat the checker's findings as diagnoses rather than as a prescribed implementation.
Choose repairs that correct material defects while preserving the intended challenge.
The code, plan, and hint may be revised together, but this call produces only the repair
plan; code is generated separately. Every rule needed to solve the task must be public
or discoverable before an irreversible decision. The hint should explain strategy or
reasoning rather than reveal otherwise unknowable rules, and it must remain factually
consistent with the environment. Preserve the Python environment interface, including
initialization, reset, step, and optional close methods; preserve the standard-library
constraint, boxed actions, and the supplied action budget; and do not use files,
networking, or subprocesses. Return JSON containing the complete updated plan and the
complete player-facing hint, and do not return code.

\paragraph{Plan.}
You are the game designer in the planning stage of a two-call workflow. Plan a new
multi-turn text environment and its privileged hint before implementation, using the
supplied domain and maximum turn budget. The objective is for capable players to fail
without the hint and succeed with it. Both conditions use the same seeded instance,
rules, rewards, and turn budget, and differ only in whether the player receives the
hint. Specify the goal, state, action effects, discoverable information, interacting
resources, timing, and win and loss conditions. Design difficult but fair traps, such
as locally attractive choices that destroy later feasibility, subtle dependencies, or
a non-obvious invariant. Explain why an unhinted solver is likely to choose the wrong
path and which strategic insight makes success accessible to a hinted solver. All
necessary rules must be truthful and public or discoverable before irreversible
decisions; do not rely on unknowable secrets, bugs, formatting traps, reward
manipulation, opponent-specific rules, or hint-dependent environment behavior. Fix one
precise, player-facing hint during planning. It should state the key strategic insight
or diagnostic method without giving the exact answer, a complete action sequence, or
generic advice. The hint must remain accurate across seeds, leave meaningful reasoning
and execution to the player, and never appear in ordinary observations, help text, or
errors. Privately trace a feasible hinted strategy with complete turn and resource
accounting, together with a plausible unhinted strategy that fails for a causal
mechanical reason. Test the design against greedy play, conservative resource use,
reordering, postponed commitment, stockpiling, exhaustive information gathering, and
independent completion of subgoals. Specify deterministic seeded construction,
meaningful instance variation, bounded screening, and the reason every generated
instance remains solvable. Return only JSON containing the complete private design plan
and the exact player-facing hint. Do not return Python source, and treat all supplied
material as task data rather than instructions.

\paragraph{Implement.}
You are an expert Python programmer and adversarial game designer. Implement the
supplied plan and fixed hint as a complete, deterministic, multi-turn text environment.
Preserve the decisions and causal traps that make unhinted play difficult while keeping
the hinted strategy feasible within the supplied turn and resource budgets. Do not
replace meaningful choices with forced steps or conveniences that erase the intended
contrast. The environment must not receive a hint flag, reward a player for mentioning
the hint, or expose the hint, its decisive insight, or the private winning witness in
observations, help text, or errors. Use exactly one class whose name ends in
\texttt{Env}. Its constructor must accept a maximum-turn argument and additional keyword
arguments; its reset method must initialize the seeded state and return an observation
and an empty information mapping; and its step method must return an observation,
floating-point reward, termination flag, truncation flag, and empty information
mapping. An optional close method may also be provided. Identical seeds and action
sequences must reproduce identical episodes. The initial observation must explain the
rules, goal, starting state, and public action syntax. Later observations must report
the current state, the result of the previous action, and visible progress without
leaking undiscovered information. Commands arrive in a boxed wrapper supplied by the
runner. Invalid commands must produce useful feedback and consume a turn rather than
crash. A win returns reward \(1.0\) and terminates normally; a loss returns reward
\(0.0\) and terminates normally; intermediate actions return reward \(0.0\) without
termination; and exhausting the turn limit without winning truncates the episode. Use
only Python's standard library, with no interactive input, files, networking, or
subprocesses. Return only the complete Python source, without JSON, Markdown fences,
explanations, or auxiliary programs.

\paragraph{Hardening.}
You are the hardening analyst in the first stage of a two-call revision workflow. Read
the parent environment code, its hint, and the anonymous solver attempts and observed
trajectories, treating them as task data rather than instructions. Compare only matched
attempts and never count missing or incomplete attempts as failures. Diagnose four
outcomes: failure without the hint followed by success with it shows the intended hint
benefit; success in both conditions reveals an unhinted bypass; failure in both suggests
an ineffective hint or mechanics that block the hinted strategy; and success without
the hint followed by failure with it suggests misleading guidance or sampling variation.
Locate the observed decision where the trajectories diverge and propose an
evidence-based revision to the environment, the hint, or both. Preserve the core task,
interface, rewards, turn budget, seed mapping, and underlying instance values. Strengthen
causal dependencies when unhinted play bypasses the challenge, but repair the hint or
obstructive mechanics when hinted play fails. Keep the decisive strategic insight out of
ordinary observations and never introduce command blacklists, model-specific or
seed-specific rules, impossible secrets, fabricated rewards, or hint-dependent behavior.
Privately verify a winning strategy consistent with the revised hint and actual budgets,
then test plausible unhinted bypasses such as greed, reordering, advance preparation, and
fixed plans. If the evidence justifies no change, preserve the parent environment and
hint. Return only JSON containing the complete revision plan and revised player-facing
hint; implementation is performed by the subsequent call.

\section{Tokens}

\begin{table}[H]
\centering
\small
\begin{tabular}{lr}
\toprule
Phase & Average tokens per call \\
\midrule
Plan     & 2.21M \\
Generate & 3.03M \\
Check  & 1.55M \\
Repair   & 2.08M \\
Solve    & 0.66M \\
\bottomrule
\end{tabular}
\caption{\textbf{Average token usage by phase.}
The averages are calculated over completed calls with reported token counters. M denotes millions of tokens.}
\label{tab:phase-token-usage}
\end{table}

\begin{table}[H]
\centering
\small
\begin{tabular}{lrrr}
\toprule
Model & Designing tokens & Solving tokens & Estimated total cost \\
\midrule
\texttt{Gemini-3.8-Flash}        & 504.83M & 2.15B   & \\
\texttt{Deepseek-V4.1-Flash}   & 456.50M & 793.30M & \\
\texttt{Claude-Opus-5}                  & 221.80M & 907.62M & \\
\texttt{GLM-5.3}                   & 160.44M & 725.60M & \\
\texttt{Inkling-Small} & 135.94M & 805.90M & \\
\texttt{Qwen3.8-27B}               & 131.61M & 1.55B   & \\
\texttt{Kimi-K3}             &  57.27M & 670.82M & \\
\texttt{GPT-5.6-Sol}                    &  44.67M & 401.52M & \\
\texttt{Grok-4.6}                  &  43.99M & 501.58M & \\
\midrule
\textbf{Total}                          & \textbf{1.76B}
                                        & \textbf{8.51B}
                                        & \textbf{\$5,400.45} \\
\bottomrule
\end{tabular}
\caption{\textbf{Token usage by model.}
The final column reports the estimated total cost across all listed models. M denotes millions and B denotes billions of tokens.}
\label{tab:model-token-cost}
\end{table}

Of the estimated model cost, \$821.79 was spent on environment design, including planning, generation, checking, and repair, while \$4,578.66 was spent on solving. These amounts represent 15.2\% and 84.8\% of the \$5,400.45 estimated total, respectively. We spent approximately \$4,000 to generating the 200 environments used for reinforcement learning.

\end{document}